\pdfoutput=1 
\documentclass{article}
\usepackage{times}
\usepackage{graphicx}
\usepackage[cmex10]{amsmath}
\DeclareGraphicsExtensions{.eps,.png,.pdf}
\graphicspath{{imglocal/}{img/}{./}}
\usepackage{caption}
\usepackage{subfigure}
\usepackage{algorithm}
\usepackage{algpseudocode}
\usepackage{amsfonts}

\begin{document}

\title{ Rounding Methods for Neural Networks with Low Resolution Synaptic Weights}
 \author{
 Lorenz K. M\"uller \&  Giacomo Indiveri\\
 Institute of Neuroinformatics\\
 University of Zurich and ETH Zurich\\
\texttt{ [lorenz/giacomo]@ini.uzh.ch}\\
 }

\maketitle
 
\begin{abstract}
Neural network algorithms simulated on standard computing platforms typically make use of high resolution weights, with floating-point notation. However, for dedicated hardware implementations of such algorithms, fixed-point synaptic weights with low resolution are preferable. The basic approach of reducing the resolution of the weights in these algorithms by standard rounding methods incurs drastic losses in performance. To reduce the resolution further,  in the extreme case even to binary weights, more advanced techniques are necessary. To this end, we propose two methods for mapping neural network algorithms with high resolution weights to corresponding algorithms that work with low resolution weights and demonstrate that their performance is substantially better than standard rounding.
We further use these methods to investigate the performance of three common neural network algorithms under fixed memory size of the weight matrix with different weight resolutions. We show that dedicated hardware systems, whose technology dictates very low weight resolutions (be they electronic or biological) could in principle implement the algorithms we study.
\end{abstract}

\section{Introduction}
\subsection{Context}
Mapping floating point algorithms to fixed point hardware is a non trivial process. The choice of mapping method can have a major impact on the performance of the fixed point system. Standard neural network algorithms typically operate on floating point parameters when simulated on conventional hardware~\cite{Ackley_etal85, Rumelhart_etal86, larochelle2011neural}. Special purpose hardware (such as FPGAs and neuromorphic chips) on the other hand commonly implement synapses with fixed point resolution and possibly a small number of bits per synaptic weight~\cite{Preissl_etal12, Chicca_etal03, Indiveri02b, Plana_etal11, merolla2014million}. How this kind of hardware can best implement neural network algorithms is an open question.

The highly related question of how biological neural networks function under limited synaptic resolution has attracted significant attention in the neuroscience community. It has been argued that limited synaptic resolution has profound effects on the learning capacity of networks that use them~\cite{ Amit_Fusi94, Fusi_etal05,lahiri2013memory, barrett2008optimal}. 
This calls into question whether it is plausible to think of the algorithms performed by biological neurons as equivalent to artificial neural nets (ANN). This analogy particularly applies to deep ANNs simulated with high resolution synaptic weights, which have been shown to be highly predictive of neural responses in visual cortex~\cite{yamins2014performance}.

We propose methods that allow artificial neural network algorithms to work with very low resolution synaptic weights using techniques from integer programming and image compression.

\subsection{Related Work}
In the computational neuroscience domain, a method for using low resolution synapses is presented in~\cite{bill2014compound}, in which a spiking neural network is trained using an STDP learning rule. However~\cite{bill2014compound} is only applicable to one specific learning rule and algorithm, a version of expectation-maximisation. In contrast we propose methods that work for several common neural network algorithms among them both discriminative and generative models. 

In the integer programming domain a method called Randomized rounding (RR)~\cite{raghavan1987randomized} has been shown to be effective in online gradient descent on the convex problem of logistic regression; in this case an upper bound on the cost introduced by RR can be given~\cite{Sculley_etal13}. We apply the same method and other methods to neural network algorithms and  also address the problem of the resolution of rounding probabilities.

A very recent paper~\cite{courbariaux2014low} examines the impact of low resolution synapses in deep learning architectures.~\cite{courbariaux2014low} focuses on different representations of low precision numbers (fixed point and floating point with different allocations of bits) with standard rounding, rather than algorithmic methods that intrinsically require lower resolution, as we do. These two approaches may well be complementary and yield best results when combined.

\subsection{This Paper}
In this paper we are interested in mapping standard neural network algorithms that use essentially continuous parameters onto equivalent ones that use low-resolution parameters. The practice of transitioning between a discrete problem and its continuous analogue, is well known in integer programming as integer relaxation~\cite{schrijver1998theory}. 

To transition from the relaxed problem (a standard neural network algorithm) to a low synaptic weight resolution version thereof, we investigate the use of two methods: one based on  randomized rounding, and the other on  a variation of an image compression technique based on k-means.
%


In particular we apply these methods to reduce the resolution of the weights in neural networks down to 2-bit resolution, while still maintaining acceptable performance figures. We show that these methods work substantially better than the naive approach based on normal rounding. This result is relevant for fixed point hardware implementations of neural network algorithms and resolves a problem in the applicability of ANNs as models of biological neural networks.

\section{Mapping Methods}
The two methods proposed for mapping continuous weight algorithms to low resolution ones differ in the way they update the synaptic weights of the neural networks: 
the first method is based on randomized rounding and works ``online'' in the sense that it changes the update procedure of the gradient descent at each update step; the 
 second method is based on k-means and is an ``offline'' method, as it compresses a learned weight matrix after training.

The benchmark that these algorithms are tested against is based on the most straight-forward technique of resolution reduction: normal rounding. For this benchmark we implemented a variant of gradient descent where at each time step the weight updates are rounded to fall onto values that are resolvable at the desired resolution. We refer to this method as \emph{online rounding}. 

\subsection{Randomized Rounding}
The first method we propose is used online, during training. It is makes use of the randomized rounding function: a  function that maps a point in a continuous one dimensional space to a point on a discrete subspace. Specifically it maps it probabilistically to either the nearest point, or the second nearest point in the discrete subspace, with a probability that is inversely proportional to the distance to the corresponding point.

\begin{algorithm}
\caption{Randomized Rounding}
\begin{algorithmic}[1]
\Procedure{RR}{$a,\epsilon$}\Comment{$a$ mapped to $\epsilon$-grid}
\State $s \gets sign(a)$
\State $p \gets \frac{|a|}{\epsilon} - \lfloor\frac{|a|}{\epsilon}\rfloor$ \Comment{probability to increase abs. val.}
\If{$p > random(0,1)$}
	\State $a \gets s \cdot \epsilon \lceil\frac{|a|}{\epsilon}\rceil$ \Comment{higher abs. val. grid point}
\Else
	\State $a \gets s \cdot \epsilon \lfloor\frac{|a|}{\epsilon}\rfloor$ \Comment{lower abs. val. grid point}
\EndIf
\State \textbf{return} $a$
\EndProcedure
\end{algorithmic}
\end{algorithm}
In the above $\lfloor \cdot \rfloor$ denotes the floor- and $\lceil \cdot \rceil$ denotes the ceiling function.

This randomized rounding method is applied during the gradient descent update that is part of all algorithms we study in this paper. The update step then looks as follows.
\begin{algorithm}
\caption{RR Gradient Descent}
\begin{algorithmic}[1]
\Procedure{update}{$\theta, d\theta, \eta, \epsilon$} \Comment{randomized rounding gradient descent, $\theta$: parameter, $d\theta$: gradient, $\eta$: learning rate, $\epsilon$: grid spacing}
\State $\theta \gets $ RR$(\theta - \eta \cdot d\theta, \epsilon)$
\State $\theta \gets clip(\theta,-1,1)$ \Comment{clip $\theta$ to allowed range}
\State \textbf{return} $\theta$
\EndProcedure
\end{algorithmic}
\end{algorithm}

We apply randomized rounding whenever a synaptic weight gets updated: Instead of being updated to a 32-bit floating point value, it gets updated to grid points $x_d \in [-1,1]$ with spacing $\epsilon$ ( i.e. $x_d \in \lbrace n\cdot \epsilon \cap [-1,1] | n \in \mathbb{N} \rbrace$). Where $\epsilon$ is chosen so that $2^i-1$ grid points are available in total. We call this the \emph{online stochastic method} with $i$ bits in the following plots.

Since in a hardware implementation the resolution of the probability in the RR procedure might be critical, we also ran this method with limited resolutions in $p$ (the resolution of $p$ was set equal to the resolution of the weights). The resolution of $p$ was reduced by standard rounding. We refer to this as the \emph{coarse p method} in the following. Notably this method does not rely on any high-resolution result.

\subsection{K-Means}
In this method we first train the neural network with high-resolution parameters, and then  use a technique taken from image compression (based on the k-means algorithm \cite{macqueen1967some}) to extract k mean weight intensities. After clustering, the value of each pixel is set to the value of the center of the cluster it belongs to. In this offline method the full weight resolution is needed during training. In principle the clustering procedure could also be applied at every step of gradient descent, which would yield an online method in some sense, but compared to RR k-means is very expensive computationally and needs `non-local' information. 

This method requires additional storage for the cluster centre values so that the memory requirement is increased by $k \cdot \log_2(p)$, where $p$ is the precision of the center value. Note that this does not scale with the matrix size $n^2$ and is negligible for $n^2 \gg k$. Since this is the regime we are interested in, we will neglect this term in the following. We will refer to this method as \emph{offline k-means}.

\section{Results and Discussion}
We applied the aforementioned mapping methods to three types of neural networks: Multi-layer perceptron (MLP)~\cite{Rumelhart_etal86}, restricted Boltzmann Machine (RBM)~\cite{Ackley_etal85} and neural autoregressive distribution encoder (NADE)~\cite{larochelle2011neural}. For all of these we investigated the impact of varying the parameter resolution under constant hidden layer size and,  for the MLP and NADE, under constant weight matrix memory (scaling the resolution by a factor of $\alpha$ also scales the size of the hidden layer by $1/\alpha$). 

The minimal resolution we consider is a 2-bit one, because these algorithms need at least three different values, a positive one, a negative one and zero. In neuromorphic hardware this can translate to two species of synpases (excitatory and inhibitory) with binary weights.

In the case of the RBM, it is difficult to give a scalar measure of performance, because the log-likelihood of some given data under a known RBM model is computationally intractable (unless the RBM is very small). To obtain a scalar measure for the performance of a generative model we applied our methods also to the NADE, an RBM inspired distribution learner of similar power, for which the log-likelihood assigned to some given data is tractable~\cite{larochelle2011neural}. To assess the performance of the RBM, samples and connection weights produced in the different conditions are plotted.

The MLP and RBM were trained on a binarized version of the MNIST hand-written digits dataset~\cite{mnist} in a theano-based~\cite{bergstra2010theano} GPU implementation of batch gradient descent. The performance measure for the MLP is the percentage of the test-set samples that were misclassified. 

The NADE was trained on the ``dna'' dataset from the libsvm webpage~\cite{chang2011libsvm} using the code provided in the supplementary materials of~\cite{larochelle2011neural} modified to allow our rounding methods. The performance measure for the NADE is the negative log-likelihood of the test set.

\begin{figure}[ht]
  \centering
    \includegraphics[width=0.6\textwidth]{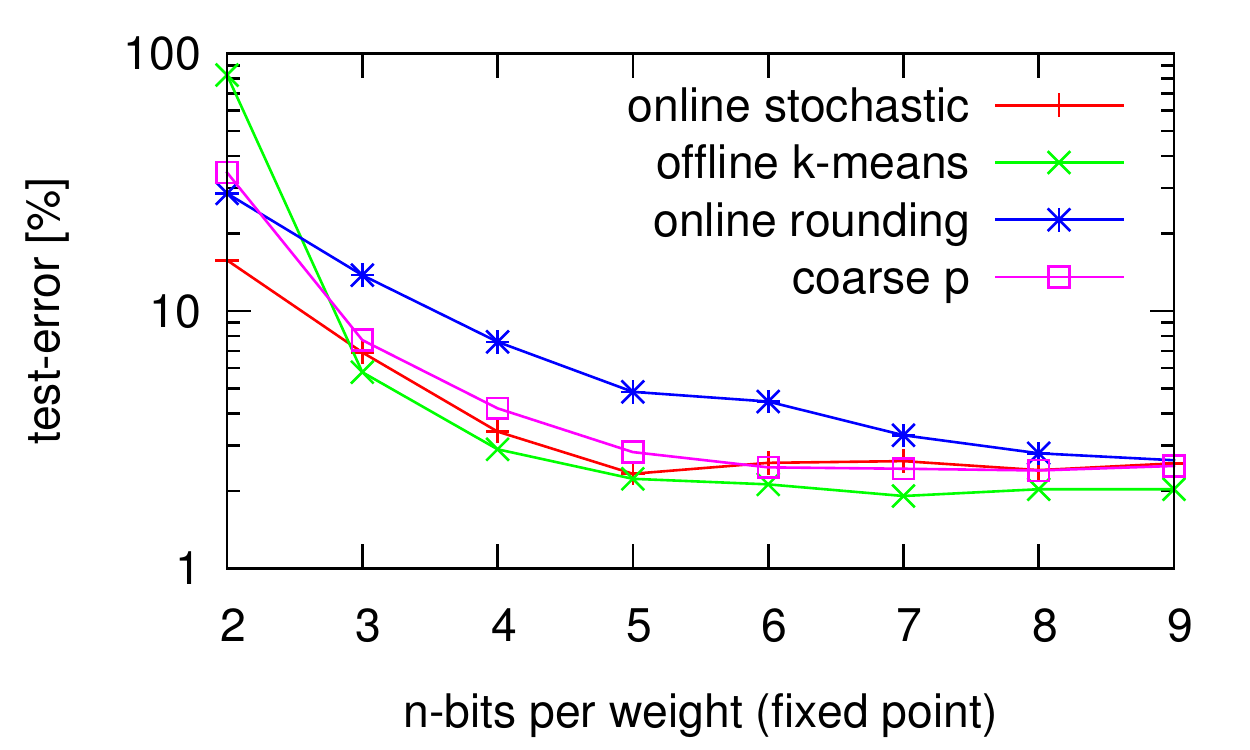}
    	\caption{The performance on MNIST of the fully trained MLP as a function of the number of bits per weight. Standard gradient descent with 32-bit floating point weights reached 1.81. There are 10 categories and the chance level is at $90\%$.}
    	  \label{fig::fixH_a}	
\end{figure}		
\begin{figure}[ht]
	\centering
    \includegraphics[width=0.6\textwidth]{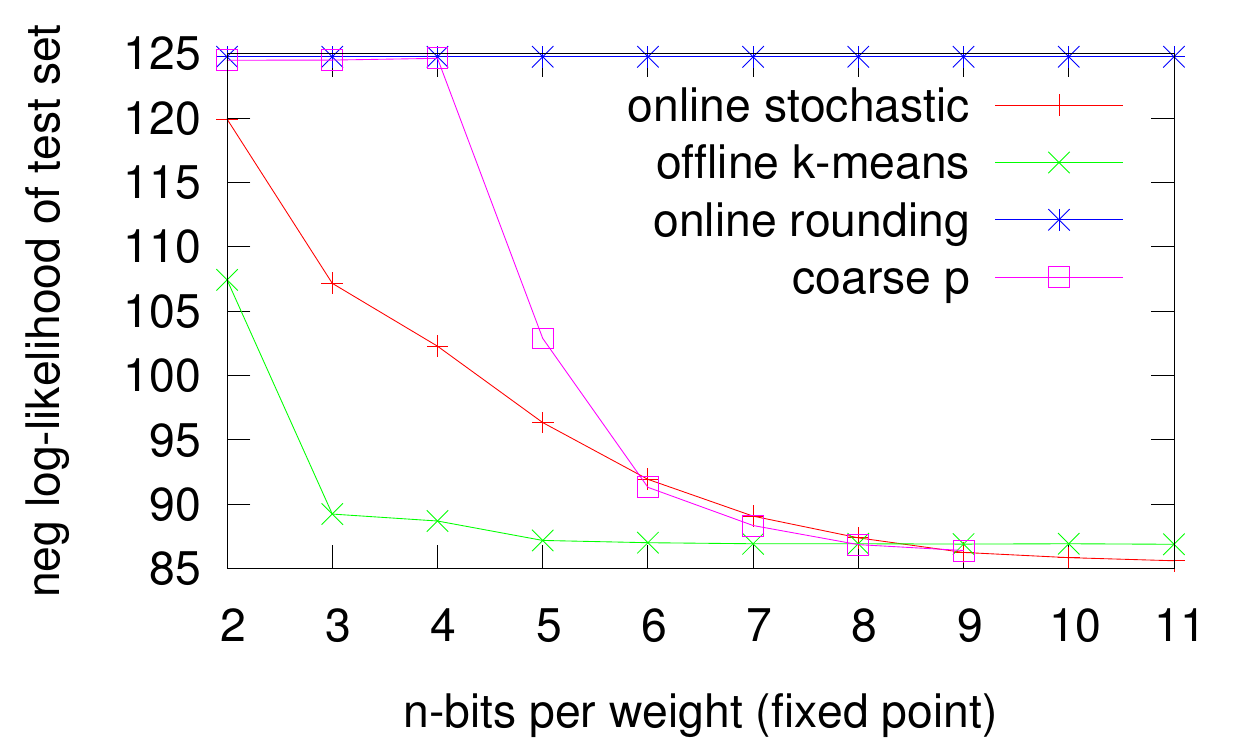}
  \caption{ The performance on the `dna' set of the fully trained NADE as a function of the number of bits per weight. Standard gradient descent with 32-bit floating point weights reached 84.6.}
      \label{fig::fixH_b}
\end{figure}	

Figures~\ref{fig::fixH_a} and~\ref{fig::fixH_b} show how the performance changes as we increase the weight resolution under fixed hidden layer size (500 units). We observe that even 2-bit weights can perform far above chance level and we see a monotonically improving performance with higher resolution and a decrease of the performance gain per added resolution bit ending in a plateau, whose floor lies near the performance of the standard gradient descent performance. The location of the plateau floor indicates a slightly poorer performance of the low resolution algorithm; this is expected, because the low resolution algorithm cannot resolve continuous parameter values so that in the end phase of the descent it will randomly jump around the minimum rather than reaching it (in the limit of infinite resolution the algorithms converge back to the high resolution algorithm and the plateau eventually reaches that level of performance).

For the `coarse p' method at 3-bit resolution performs similarly well as the 6-bit normal rounding method that uses equally much memory per weight update. In contrast for the NADE the normal rounding method at 10-bit resolution peformed at chance level, while the 5-bit `coarse p' performed above chance. 
%



\begin{figure}[ht]
  \centering
    \includegraphics[width=0.6\textwidth]{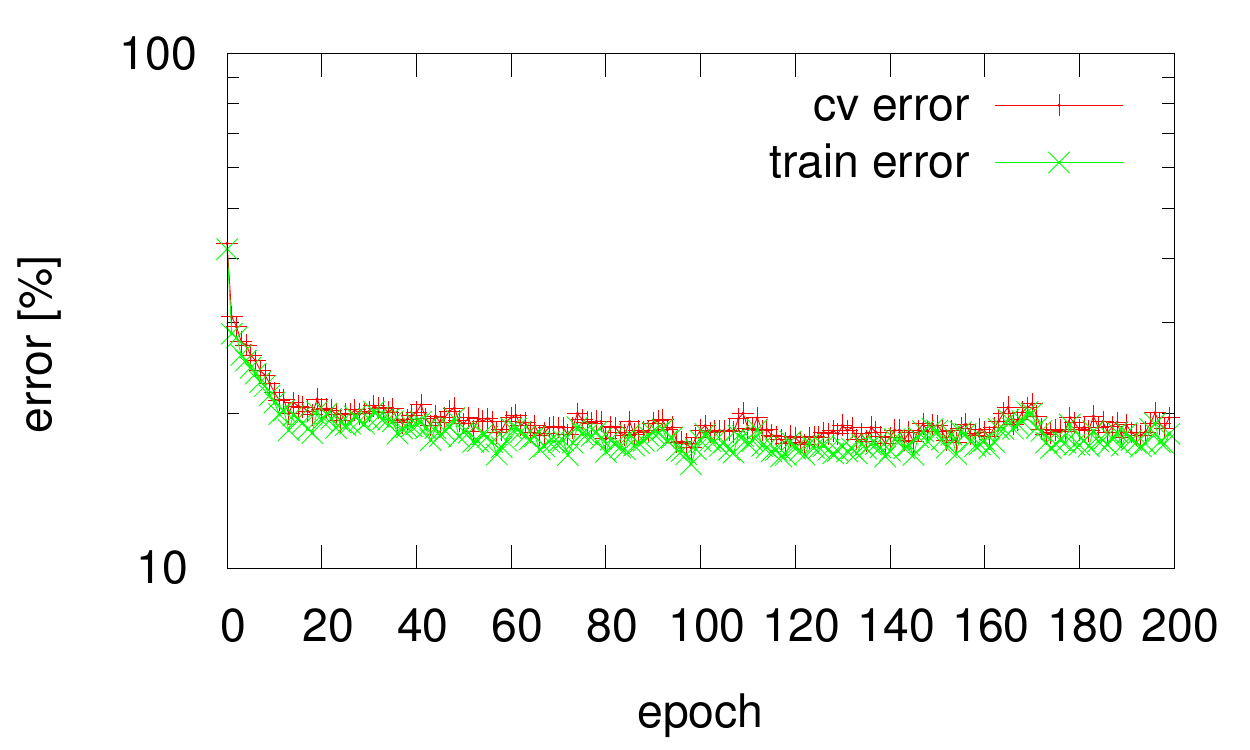}
  \caption{The cross-validation and training error are very close for a 2-bit resolution MLP on MNIST. This indicates a high-bias model.}
      \label{fig::learning_curves}
\end{figure}	

The learning curves for a low resolution MLP (500 hidden units, 2-bit resolution) in Figure \ref{fig::learning_curves} show that for very low resolutions the model performs very similarly on the training as on the cross-validation set. This indicates that this model is limited by its expressive power, rather than by the learning algorithm (it has `high bias' rather than `high variance'). In light of this randomized rounding can also be interpreted as a regularization procedure.

\begin{figure}[ht]
	\centering
    \includegraphics[width=0.6\textwidth]{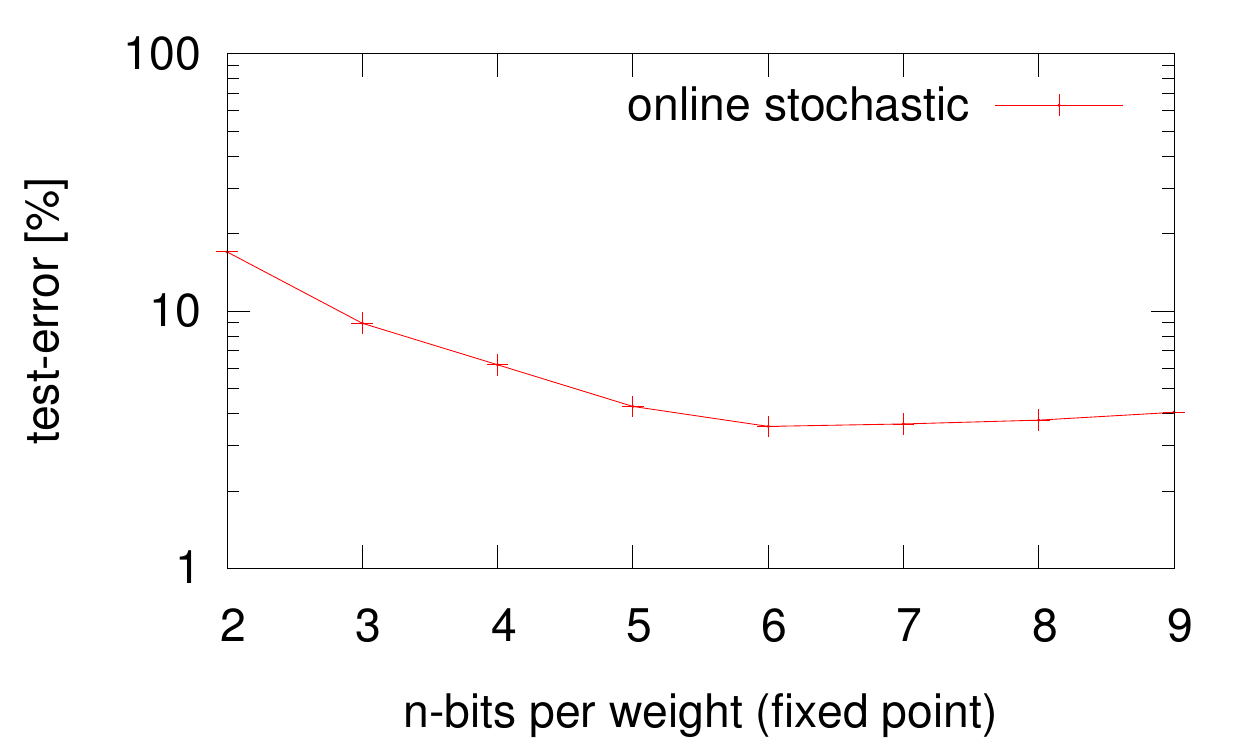}
  \caption{ The performance on the `dna' set of the fully trained MLP as a function of the number of bits per weight  while keeping the matrix memory size constant. The minimum lies at 6.}
      \label{fig::fixM_b}
\end{figure}	 
\begin{figure}[ht]
  \centering
    \includegraphics[width=0.6\textwidth]{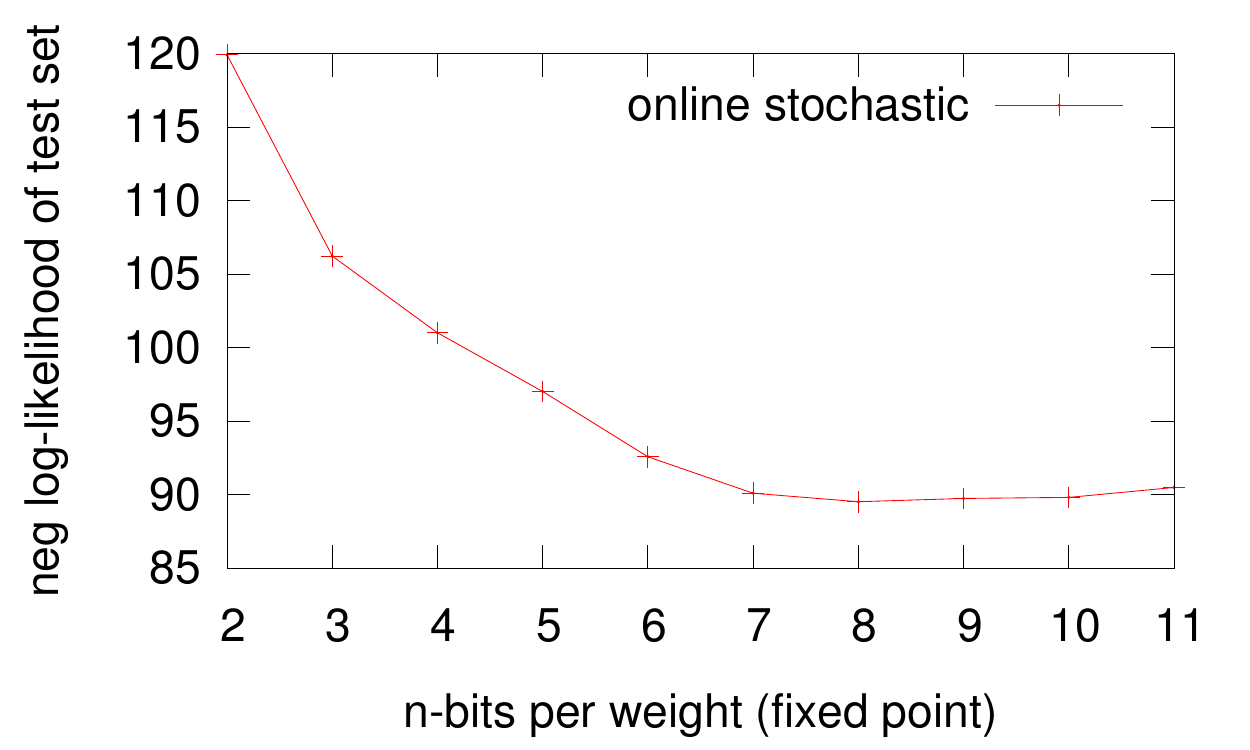}
    \caption{The performance on MNIST of the fully trained NADE as a function of the number of bits per weight while keeping the matrix memory size constant. The minimum lies at 8.}
        \label{fig::fixM_a}
\end{figure}

Figures~\ref{fig::fixM_b} and~\ref{fig::fixM_a} show how the performances of NADE and MLP change as we increase the weight resolution while keeping the memory size of the weight matrix fixed at 400 bits. Under these conditions it is clearly preferable to choose an intermediate resolution.

\begin{figure}
\centering
\begin{subfigure}{}
	\includegraphics[width=0.08\textwidth]{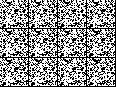}
	~
	\includegraphics[width=0.08\textwidth]{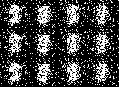}
	~
	\includegraphics[width=0.08\textwidth]{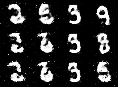}
	~
	\includegraphics[width=0.08\textwidth]{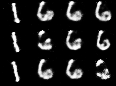}
\end{subfigure}
\hspace{2cm}
\begin{subfigure}{}
	\includegraphics[width=0.08\textwidth]{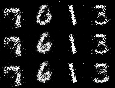}
	~
	\includegraphics[width=0.08\textwidth]{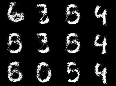}
	~
	\includegraphics[width=0.08\textwidth]{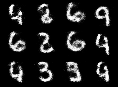}
	~
	\includegraphics[width=0.08\textwidth]{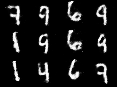}
	
\end{subfigure}
{\small(a) online rounding \hspace{5.5cm} (b) online stochastic}
\\ 
\begin{subfigure}{}
	\includegraphics[width=0.08\textwidth]{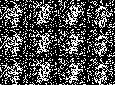}
	~
	\includegraphics[width=0.08\textwidth]{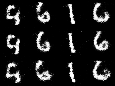}
	~
	\includegraphics[width=0.08\textwidth]{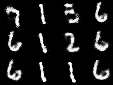}
	~
	\includegraphics[width=0.08\textwidth]{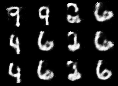}
\end{subfigure}
\hspace{2cm}
\begin{subfigure}{}
	\includegraphics[width=0.08\textwidth]{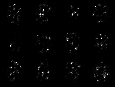}
	~
	\includegraphics[width=0.08\textwidth]{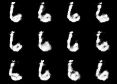}
	~
	\includegraphics[width=0.08\textwidth]{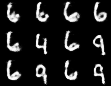}
	~
	\includegraphics[width=0.08\textwidth]{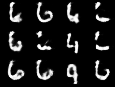}
\end{subfigure}
\\
{\small(c) coarse p \hspace{6.2cm} (d) offline k-means}
\caption{Samples (activations, not binarized) for 2, 4, 6, and 8-bit RBM trained with the four different resolution reduction methods. Inside each picture: Four different initial conditions (random test sample) horizontally arranged, over 3000 passes, printed every 1000 steps vertically arranged.}
\label{fig::rbm_samp}
\end{figure}

\begin{figure}
\centering
\begin{subfigure}{}
	\includegraphics[width=0.08\textwidth]{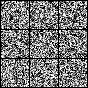}
	~
	\includegraphics[width=0.08\textwidth]{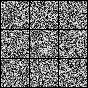}
	~
	\includegraphics[width=0.08\textwidth]{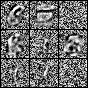}
	~
	\includegraphics[width=0.08\textwidth]{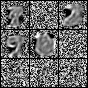}
\end{subfigure}
\hspace{2cm}
\begin{subfigure}{}
	\includegraphics[width=0.08\textwidth]{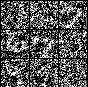}
	~
	\includegraphics[width=0.08\textwidth]{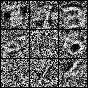}
	~
	\includegraphics[width=0.08\textwidth]{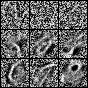}
	~
	\includegraphics[width=0.08\textwidth]{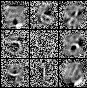}
	
\end{subfigure}
{\small(a) online rounding \hspace{5.5cm} (b) online stochastic}
\\ 
\begin{subfigure}{}
	\includegraphics[width=0.08\textwidth]{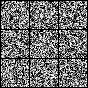}
	~
	\includegraphics[width=0.08\textwidth]{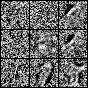}
	~
	\includegraphics[width=0.08\textwidth]{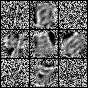}
	~
	\includegraphics[width=0.08\textwidth]{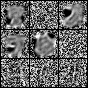}
\end{subfigure}
\hspace{2cm}
\begin{subfigure}{}
	\includegraphics[width=0.08\textwidth]{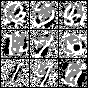}
	~
	\includegraphics[width=0.08\textwidth]{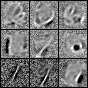}
	~
	\includegraphics[width=0.08\textwidth]{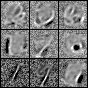}
	~
	\includegraphics[width=0.08\textwidth]{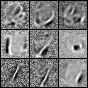}

\end{subfigure}
{\small(c) coarse p \hspace{6.2cm} (d) offline k-means}
\caption{Final receptive fields for 2, 4, 6, and 8-bit RBM}
\label{fig::filters}
\end{figure}

Figure \ref{fig::rbm_samp} shows activation probabilities for samples given by RBMs with different weight-resolutions (all have hidden layer size 500) trained with PCD-15 \cite{tieleman2008training}. As with the other algorithms the quality improves with higher resolutions, but even 2-bit weights already result in clearly recognisable digits (albeit noisy ones) for the randomized rounding method.

Figure \ref{fig::filters} shows receptive fields learned in RBMs with varying weight resolutions. Notably there are some hidden units whose receptive fields `look' very noisy for low resolution weights. However, it may well be the case that it is difficult to judge by eye what constitutes a `useful' receptive field; conversely the weights for the 2-bit k-means method `look' useful but do not produce good samples.

A particularly interesting application of randomized rounding gradient descent, would be a neuromorphic neural network implementation with memristive synpapses that exhibit probabilistic switching~\cite{medeiros2011lognormal}. For other algorithms it has already been proposed that this behaviour could be exploited in neuromorphic hardware~\cite{bill2014compound}. Thus it could be possible to implement the randomized rounding step directly in the memory unit, without need for a random number generator. 

\section{Conclusion}
We presented two methods to reduce the resolution of common neural network algorithms to very low resolutions, while maintaining comparatively good performance: Randomized rounding, an `online' method for performing gradient descent with low resolution parameters and K-means rounding, a post-processing method that  reduces the resolution of the weights in  neural networks trained with normal high-resolution parameters. We applied these methods on the  MLP, NADE and RBM neural network algorithms, showing  a graceful degradation of performance with decreasing weight resolution.
 
Using these techniques, the performance of the algorithms plateaued around the 10-bit resolution mark for datasets and parameter ranges we studied; no substantial improvement was made above this resolution and additional memory was better invested in larger hidden layers. The offline method based on k-means produced better results than the online method, and both performed substantially better than rounding.

Overall we find that there are no fundamental problems with the use of even binary excitatory and inhibitory synapses (i.e. 2-bit weights) in the tested ANN algorithms. Such low resolution synapses are common in neuromorphic hardware and 2-bit weights is the lower bound for the resolution of biological synapses. 

Increasing to 6 or 8-bit resolution yielded substantial performance improvements with RR gradient descent. At very low resolutions it seems sensible to forgo learning on a dedicated hardware implementation, if it is not inherently required; then a system of the same memory size can deliver a better performance using an offline compression of the weight matrix. 

Notably randomized rounding worked well as a mapping method to fixed point weights for all algorithms we tested and down to very low resolutions. We speculate that other gradient-descent-based algorithms may well be similarly compatible with randomized rounding. 
\section*{Acknowledgements}
This work was supported by the European Commission under the ``neuroP''  Project, ERC-2010-StG-257219-neuroP.

\bibliographystyle{unsrt}
\bibliography{biblio}

\end{document}